# SRGAN: Training Dataset Matters

**Nao Takano**

nao.takano@ucdenver.edu
Computer Science and Engineering
University of Colorado Denver

**Gita Alaghband**

gita.alaghband@ucdenver.edu
Computer Science and Engineering
University of Colorado Denver

## Abstract

Generative Adversarial Networks (GANs) in supervised settings can generate photo-realistic corresponding output from low-definition input (SRGAN). Using the architecture presented in the SRGAN original paper [2], we explore how selecting a dataset affects the outcome by using three different datasets to see that SRGAN fundamentally learns objects, with their shape, color, and texture, and redraws them in the output rather than merely attempting to sharpen edges. This is further underscored with our demonstration that once the network learns the images of the dataset, it can generate a photo-like image with even a slight hint of what it might look like for the original from a very blurry edged sketch. Given a set of inference images, the network trained with the same dataset results in a better outcome over the one trained with arbitrary set of images, and we report its significance numerically with Fréchet Inception Distance score [22].
keywords: GAN, SRGAN, FID

## 1. Introduction

Generative Adversarial Networks are a type of deep neural networks that are used to generate images by using two networks; the generator and the discriminator. The generator attempts to generate images from an array of often smaller size, and the discriminator determines whether the generated images are real or fake by finding the Nash equilibrium of a game [[7], [8]]. GANs have shown to be very effective methods for Super Resolution imaging (SR), also referred to as Single image super-resolution (SISR), which is a class of techniques that enhances the resolution of an imaging system. It aims to refine and convert an image of low-resolution to that of high-resolution. Following the work by Dong et al. [25], Ledig et al. [2] demonstrated that using the Generative Adversarial Network as well as pre-trained VGG network [26] improve the performance of SR (SRGAN).

SRGAN has achieved generating high resolution images by training rather general image sets, i.e., any arbitrary image set can be used for training. But if one is to enhance a particular type of image, such as that of human face, a building, or furniture, should the network be trained with the given specific features that will appear in the inference images? The question of how the selection of training data affects the SRGAN has not been well studied.

This paper explores effective ways of training to achieve super-resolution and other applications, using various datasets in SRGAN. In order to answer a question such as *"If we want to reconstruct high-resolution face images from low-resolution ones using SRGAN, should we train it with the dataset of face images, or does it matter what kind of images we train?"*, we use three different datasets for training and measure the effects on the resulting images. In other words, this study attempts to uncover how SRGAN uses neural networks to generate super-resolution images by learning other similar images.

Our main contributions are as follows:

- With FID, we show numerically that for the best result one should use the same image-type dataset that will be used at inference time.
- Demonstrate examples of other usages of SRGAN, such as Coloring and Edges to Photo, as applications of image-to-image translation.

After Section 2, Related Work, we describe the dataset and the network architecture, along with the experimental methods in Section 3. Section 4 describes the results of the experiments in Section 3. In Section 5, we explore usages of SRGAN beyond super-resolution, coloring and edges to photo, and the paper concludes with Section 6.

## 2. Related Work

### 2.1 Unsupervised Training

With Generative Adversarial Networks, along with Variational AutoEncoders (VAE) [20] and other techniques, a random noise is often used as an input to draw images in an unsupervised manner. Since its inception in 2014 [7], GAN has evolved within this domain for generating clearer images and more stable training, with the input of a sample from a Gaussian distribution $z$. Radford et al. [10] successfully generated convincing images by using a convolutional neural network that has no fully connected layers and replaced max-pooling with strided convolutions. Arjovsky et al. [11] demonstrated a stable training by measuring divergence with Earth Mover (EM) distance, a distance between two probability distributions that is to be minimized continuously. It was followed by Gulrajani et al. [12], which proposes weight regularization rather than weight clipping. Mao et al. [13] introduced a least-squares loss for slower saturation of the loss function. And Berthelot et al. [14] proposed a loss derived from the Wasserstein distance with an auto-encoder based training as a discriminator. Meanwhile, with increasing members of the GAN family, efforts have been made to organize, sort, and evaluate different flavors of GANs [18].

### 2.2 Supervised Training

On the other hand, attempts to control the output more precisely by way of supervised training settings took place. The training is conducted by feeding an image to the network as an input (as opposed to the random noise z), and form a one-to-one mapping with the corresponding output image. Our work focuses on this area. SRGAN [[2]] is one of the prominent works of one-to-one mapping.

Conditional GAN (cGAN) [[15], [16], [17]] inserts a condition in the network and tries to generate an output with specific attributes. Notably, in close relation to our work, Isola et al. [3] showed that cGANs can achieve image-to-image translation. In this regard, SRGAN is a part of the broader term of Conditional GANs.

The supervised nature of these GANs requires paired training data, which is often difficult to obtain except in the case where one can generate input images from the original ones, as demonstrated in this paper. CycleGAN [9] eliminates this requirement; it can translate images in one domain to another without pairs of training data.

## 3. Methods and Experimental Settings

### 3.1 Dataset

We use three different datasets:

1. CelebA for faces – we use the "aligned" version, which has been compiled from the original version of "in the wild". Every image is a uniform size of 178 x 218 and contains all peoples' faces in the center of the frame. (Total number of images: 202,599) [27]
2. Lsun Dining Room – consisting of dining tables, chairs, etc., representing indoor furniture (657,571) [28]
3. Lsun Tower – architectural buildings in the outdoors (708,264) [28]

The idea of selecting these three datasets is to contrast them to each other, where each dataset representing unique characteristics of images of a similar kind (faces, towers, furnished rooms). We created three different pre-trained networks, corresponding to respective datasets. When the trainings are completed, each of which should have learned human faces, indoor furniture, and buildings outdoor, respectively. For our experiment, a single training consists of 200,000 images, which are selected randomly from one of the datasets and run for 10 epochs.

Note that the aligned CelebA contains faces in the frame consistent in all the images in such a way that we can simply center-crop and resize them in the same ratio, while Lsun images are taken from the wild, and objects are placed randomly in an image. Thus, naturally the network learns faces more easily than dining rooms or towers. All the images are resized (CelebA) or randomly cropped (Dining Room and Tower) to be 128×128 in size.

### 3.2 Network architecture

#### 3.2.1 Architecture

We used code from Github [[1]] which closely presents the original SRGAN [[2]] and made very few changes to the network itself. We used different datasets for comparison. The generator of the network has eight residual blocks (the original SRGAN uses 16 blocks), each of which consists of Conv, BN, and Parametric ReLU, followed by layers of PixelShufflers, which are inserted to accommodate various upscale ratios (The original SRGAN has two PixelShufflers to facilitate 4×4 upscaling factors). The discriminator also contains eight convolutional layers, consisting of Conv, BN, and Leaky ReLU, but with more feature maps than those of the generator. Both networks use the kernel size 3 and no max-pooling is used. In order to properly measure the effects of the choice of datasets, we used the fixed network architecture throughout the experiments. (Figure 1)

Although the study of the network construction is beyond the scope of our work, a few points should be made. There are discussions as to which network between the generator and discriminator should be more powerful, such as how fast the generator learns than the discriminator does, or vice versa. Despite suggestions that the parameters of the discriminator should be updated more frequently than the generator, many implementations use the same frequency of parameter updates between the two networks. We follow the latter. In terms of the network size, the discriminator in our network is several times larger than its counter-part.

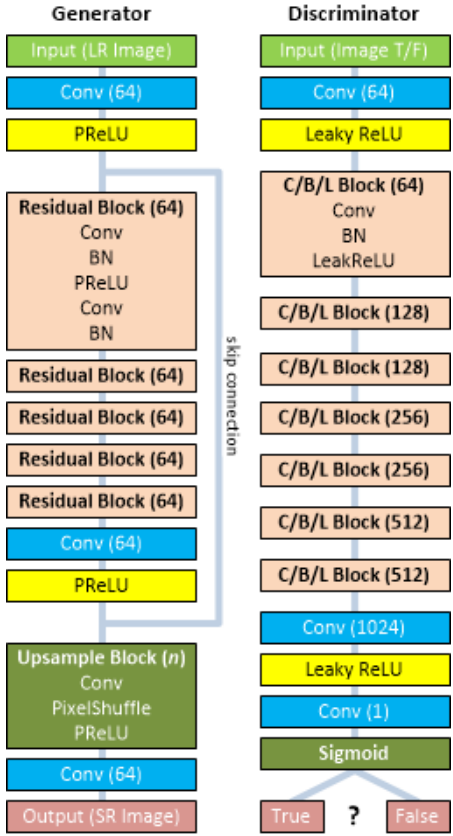

**Figure 1**: Architecture of the network. The numbers in the parenthesis indicate the number of feature maps.

### 3.2.2 Loss function

With a pair of training images, the discriminator calculates the loss function – pixel-wise distances between the target image and the input image. This is a straight forward way to minimize the loss in image-to-image translation. The original SRGAN paper [2] uses Mean Squared Error (MSE, or L2) loss in its content loss calculation. While we also used L2 loss for the majority of experiments, in Section 5 we used both L1 and L2 losses for comparison. In general, the L2 function performs well as long as outliers (noise in the image) are not present in the input.

$$L_{L1}(G) = \sum_{x=1}^{W}\sum_{y=1}^{H} \left\| I_{x,y}^{target} - G(I^{input})_{x,y} \right\|$$
$$L_{L2}(G) = \sum_{x=1}^{W}\sum_{y=1}^{H} \left( I_{x,y}^{target} - G(I^{input})_{x,y} \right)^2$$

As part of the content loss, 16 layer VGG network [26] was used throughout the experiments along with L1 or L2 loss.

### 3.2.3 Batch Normalization

Batch Normalization (BN) is a normalization technique used on the input of each layer to reduce covariance shift (changing of distribution of activations in intermediate layers) [[5]]. Wang, et al. [[4]] claims that removing the BN improves the image quality, but it is also known to allow a wider range of the selections for hyper-parameters. Determining hyper-parameters increases the experimental budget exponentially when we change the image size, and BN helps alleviate this cost. Using Batch Normalization helps the network to converge more easily.

The network adopted in this paper uses Batch Normalization at each layer in all experiments.

## 4. Super-Resolution with 4×4 up-scaling images

### 4.1 Method

The up-scale ratio we use in this section is 4×4, as is typically done in SRGAN. We create three trained-networks by training three datasets, namely, CelebA [27], Lsun Dining Room [28], and Lsun Tower [28]. Given a target image of 128×128, an input image is created by resizing the target image by 4×4 with bi-cubic interpolation, resulting in a 32×32 low resolution (LR) image. This input is then passed to the generator, resulting in an image with the original size of 128×128, yet still low resolution, at which point we have a pair of images; one is the original HR image (target), the other is the LR output by the generator, both the same size. Now, the loss is calculated between the pair which is to be minimized during the training. This process is repeated with 200,000 different pairs, consisting of one epoch.

We trained 10 epochs for each of three networks. With Nvidia GTX 1080 Ti, using a single GPU it takes approximately one hour for an epoch, thus about 10 hours to finish training. After creating the three trained networks, each with trained with faces, dining rooms, and towers, we then test each for inference with the test sets of the three datasets of face, dining room and tower, totaling 9 tests.

### 4.1.1 Evaluation of the performance metrics

The image-to-image translation involves the supervised training of two images; original or target, and the corresponding image generated from the LR input, to be measured its fidelity to the original. Thus, direct comparison of two images between the original and the generated image would make sense. We can measure the Euclidean distance, often as MSE, of every pixel between pairs of the two images, then average them over the entire test set. Peak

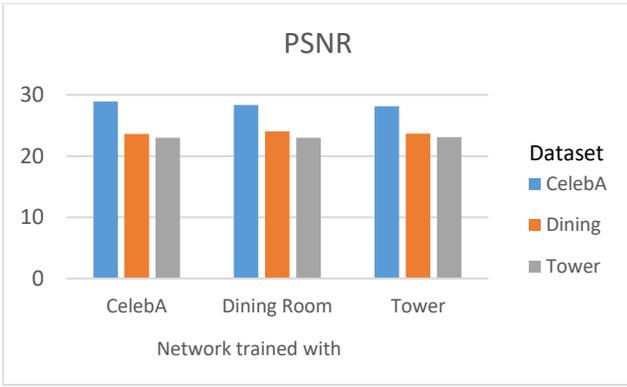

**Chart 1**: Pixel-wise Euclidean Distance Metrics – PSNR. The higher the value, the better.

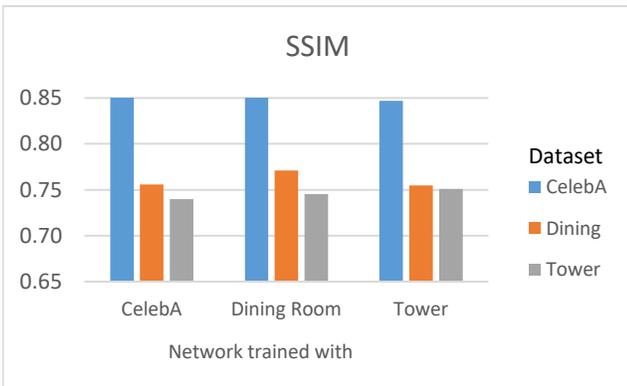

**Chart 2**: Pixel-wise Euclidean Distance Metrics – SSIM. The higher the value, the better.

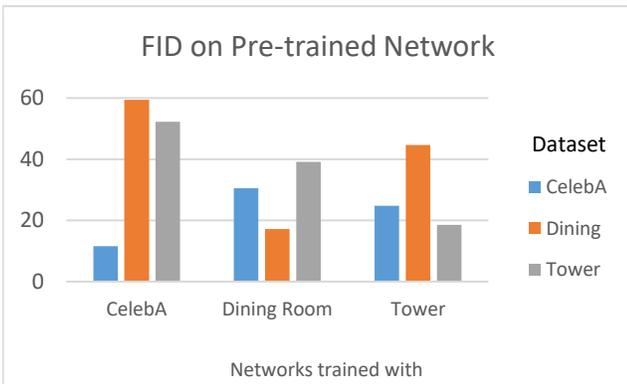

**Chart 3**: Each pre-trained network that has been trained with CelebA, Dining Room, and Tower is tested against three datasets. The lower the value is, the better.

Signal to Noise Ratio (PSNR) and Structural Similarity Index (SSIM) are derivatives of MSE metrics. PSNR represents a measure of the peak error (noise). The lower the value, the lower the error. SSIM gives a value of 1.0 if the two images are identical. The lower the SSIM value, the bigger the difference. Traditionally they have been used to measure image qualities in the reconstruction of image compression, however, for GANs those measurements do not represent the true quality of the generated output [[2], [4], [6], [8], [18], [21], [22]]. The reason they do not represent visual perception is well described in Theis et al. [21]. Even for a pair of identical images, a shift by only one pixel on one of the images causes a large increase in the distance of the metrics. Nevertheless, Chart 1 and 2 show these MSE-based quantitative results and will be compared to Fréchet Inception Distance (FID) [22], which corresponds to more realistic qualitative measure

Fréchet Inception Distance uses a pre-trained Inception network [24] and calculates Fréchet distance between two multivariate Gaussian distributions with mean $\mu$ and covariance $\Sigma$,

$$\mathrm{FID}(x, g) = \|\mu_x - \mu_g\|^2 + \mathrm{Tr}\left(\Sigma_x + \Sigma_g - 2(\Sigma_x \Sigma_g)^{1/2}\right)$$

where $x$, $g$ are the activations of the pool_3 layer of the Inception-v3 net for real samples and generated samples, respectively. We take 2,599 images from test set, and generate the corresponding inference images through one of our pre-trained networks, resulting in two sets that are to be measured by the FID calculator [23]; 2,599 test images and 2,599 output images. Note that the number of images comes from the CelebA dataset, whose total is 202,599, of which we used 200,000 for training, thus leaving 2,599 for testing. The sample size must be consistent throughout the measurements, thus we randomly select 2,599 images from each of test sets of Dining Room and Tower as well.

Measuring FID appropriately shows that an image of CelebA, Dining Room, or Tower is best inferred through the network that has been trained with CelebA, Dining Room, or Tower, respectively, as seen in the [Chart 3]. FID measures can better support what we observe visually in Figure 2 as described in the next section.

### 4.1.2. Visual Analysis of the outputs

More obvious differences can be observed in actual images shown in Figure 2. For example, the top three images are CelebA images and we can see that the networks trained with Dining Room and Tower never learned the shape of eyes, whereas the network that has been trained with faces clearly shows pupil/iris (dark area in the center of an eye) and sclera (white area surrounding iris).

The 4[th] and 5[th] rows are Dining Room images. In the 4[th] row, the edge of the white table is more clearly shown, inferred by the network trained with Dining Room. Similarly, the image in the 5[th] row shows columns of the chair (vertical wooden back support) more clearly through the network trained with it. On the other hand, in the images with the network trained with Tower, one can observe bricks and window-like texture on the surface of the buildings more clearly (rows 6 and 7).

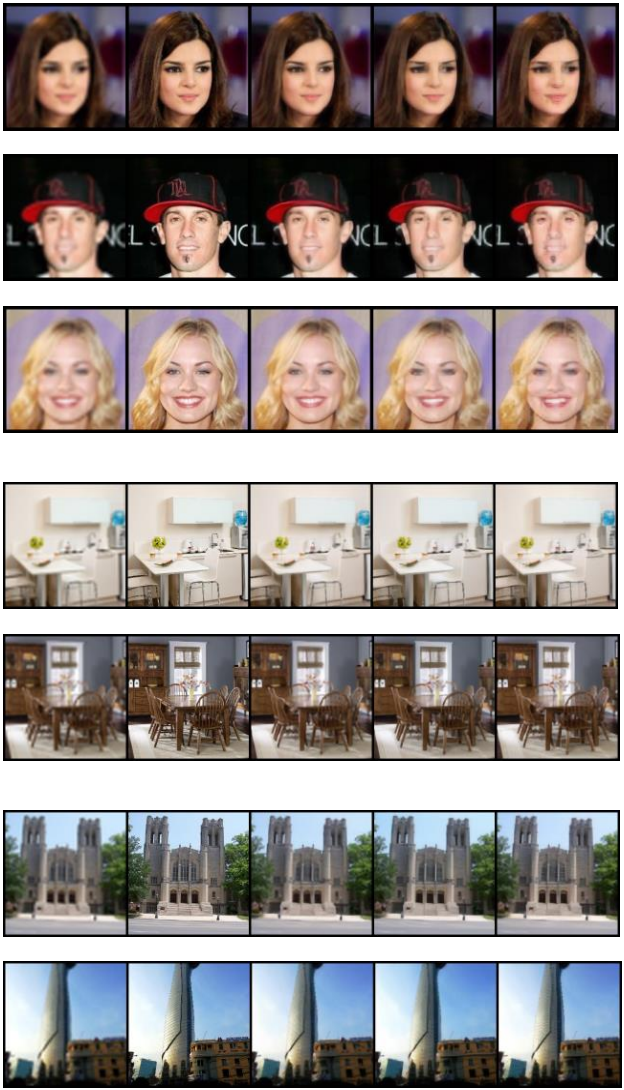

**Figure 2**: Top 3 rows: CelebA images. Center 2 rows: Dining Room. Bottom 2 rows: Tower.
From left to right: Input LR image, Target Image,
Output inference by pre-trained network with CelebA,
Output inference by pre-trained network with Dining Room,
Output inference by pre-trained network with Tower.

## 5. Coloring and Edges to Photo

### 5.1 Coloring

There have been various methods of coloring single channel images – i.e., grayscale and edged. But CNN architectures, along with Transposed/Fractionally-Strided Convolutions (or deconvolutions), have been used more recently [[3], [19]]. We used this image-to-image translation with L2 loss function for the colorization with the same architecture of the network used for SRGAN. We used upscale ratio 1 (No blurring operation in the inputs) for all the trainings.

For inference, we only used corresponding pre-training networks; i.e. for the inference of face images, we used the network trained by face dataset only, for dining room images we used the network trained by dining room dataset only. Before the training, we prepared the input data with gray-scale images using OpenCV.

#### 5.1.1. Face

Figure 4 shows the result of our experiment for face coloring. As can be seen, regardless of race the difference in skin color of human faces is relatively small. However in the testing, the network properly redraws the original skin color in the output images. Note that the network never learned the color of the background, and more generic beige color is used for it (3$^{rd}$ and 4$^{th}$ rows on Figure 4).

#### 5.1.2. Dining Room

Figure 5 shows the result of our experiment for dining room coloring. Many training images contain tables and chairs, and very often if the color of the table is white, the chairs are brown, and vice versa. Two examples of those color combinations are presented. It appears the network detects the texture of materials to determine the color of furniture.

#### 5.1.3. Tower

As shown in Figure 6, the buildings in the dataset do not exhibit a common color, but other elements, such as the sky or illumination of the building at night, will be regenerated in the output, which indicates that the network picks up colors of the most common denominator found in the training images.

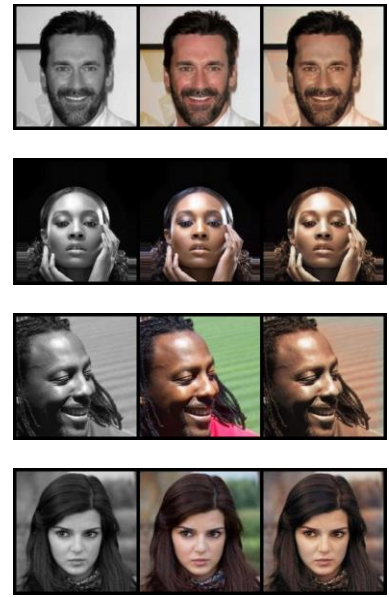

**Figure 4:** Converting Gray-Scale to Color (Face)
Left: black and white, Center: original, Right: output.

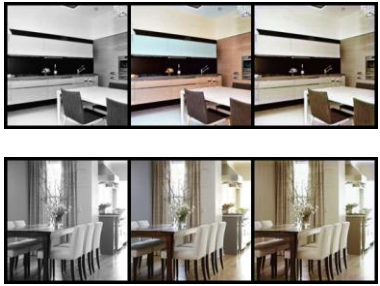

**Figure 5:** Converting Gray-Scale to Color (Dining Room)
Left: black and white, Center: original, Right: output.

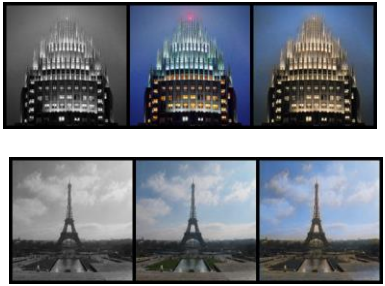

**Figure 6:** Converting Gray-Scale to Color (Tower)
Left: black and white, Center: original, Right: output.

### 5.2 Edges to Photo

In this section, we conducted additional experiments with two loss functions described in Section 3.2.2, as well as several upscale values, to see how blurry edge images can generate corresponding colored images. We were able to generate images only for CelebA; the network did not converge for Dining Room or Tower in training. In the CelebA datasets, the objects (faces) are placed roughly at the center of the frames, but Lsun images contain multiple objects "in the wild", not aligned in any way as in CelebA, blurring a focal point to learn a single object. We used OpenCV to generate edged images with its Canny edge detector before training.

Setting the upscale value 1 (input and target images have the same size) induces artifacts, although the output image has better resolution than that of the upscale value of 2 or greater. As we blur the input, the artifacts disappear in the output and smooths out the details. In the very blurry output, such as ×8 and ×16, the network still draws human faces, indicating what it has learned by the training. In terms of the loss functions L1 or L2, we observe clear differences, but it is hard to say which one is better than the other for all cases (Figure 7).

The way the artifacts show up in the output draws an attention. Our experimental results and observations suggest that in most cases, either experimental or practical, the objective of the type of exercise presented in this section is a general restoration of the original object, per se, so the features in objects in the input should be in such a way that they represent a general guideline, like DNA, to represent the original. Neural network of SRGAN and Image-to-Image translation learns to capture these features, it seems, in the form of the guideline to reproduce the original image. For a pre-trained network, the more articulate the guideline is, the better it produces the image. However, if the guideline in the input is overly articulated, there will be artifacts in the output. The greater the detail the input is depicted, the more amplified the artifacts will be, in which case it appears the output fails to generalize the original object.

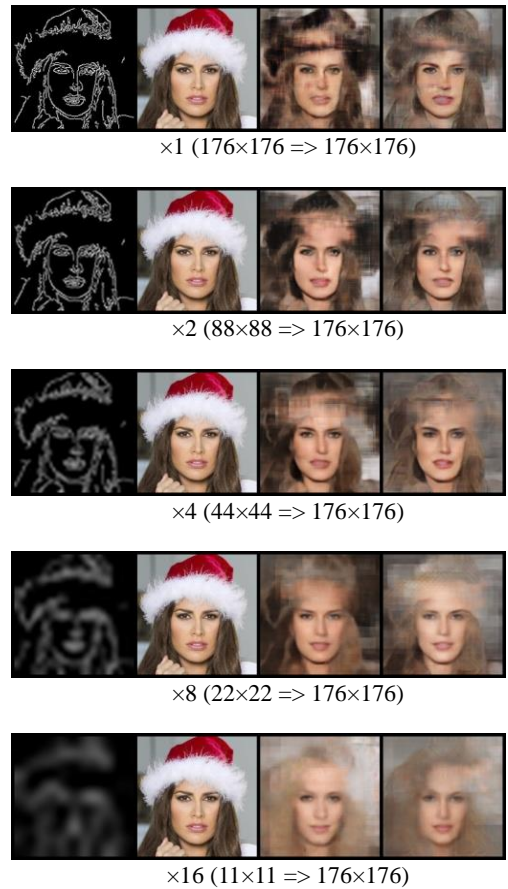

×1 (176×176 => 176×176)

×2 (88×88 => 176×176)

×4 (44×44 => 176×176)

×8 (22×22 => 176×176)

×16 (11×11 => 176×176)

Figure 7: Converting Edges to Photo
From left to right: Input, Target, Output L1, and Output L2

### 6. Conclusion

In all the experiments conducted in this paper, we used a fixed network architecture of the GANs. We showed that given a set of inference images, the network trained with the same dataset results in a better outcome. This is numerically presented in FID scores. SRGAN fundamentally learns objects, with their shape, color, and texture, and redraws them in the output rather than attempting to sharpen edges.

We also showed that one-to-one mapping of image translation inherent in supervised training can be used in coloring and converting edges to photo. Once the network learns objects, it can regenerate them from a very faint sketch that suggests the original. The question remains as to whether the result presented here is consistent with all other architectures of the network. We further need to investigate the artifacts presented in the previous section in future work.